\documentclass[letterpaper]{article} 
\usepackage{aaai24}
\usepackage{times}  
\usepackage{helvet}  
\usepackage{courier}  
\usepackage[hyphens]{url}  
\usepackage{graphicx} 
\usepackage{natbib}  
\usepackage{caption} 
\usepackage{bm}
\usepackage{amsmath,amssymb,amsthm}
\usepackage{mathtools}
\usepackage{cleveref}
\usepackage{autonum}
\usepackage{cite}
\usepackage{multirow}
\usepackage{booktabs}

\usepackage{algorithm}
\usepackage{algpseudocode}

\newcommand{\argmin}{\mathop{\rm argmin}\limits}
\newcommand{\vect}[1]{\mathbf{#1}}
\newcommand{\balpha}{\boldsymbol\alpha}
\newcommand{\bbeta}{\boldsymbol\beta}

\DeclareMathOperator{\EE}{\mathbb{E}}

\DeclarePairedDelimiter\floor{\lfloor}{\rfloor}
\DeclarePairedDelimiterX\inner[2]{\langle}{\rangle}{{#1},{#2}}
\DeclarePairedDelimiter\abs{|}{|}
\DeclarePairedDelimiter\norm{\|}{\|}

\DeclarePairedDelimiter\set{\{}{\}}
\DeclarePairedDelimiter\prn{(}{)}
\DeclarePairedDelimiter\bra{[}{]}
\DeclarePairedDelimiterX\Set[2]{\{}{\}}{\mspace{2mu}{#1}\;\delimsize|\;{#2}\mspace{2mu}}
\DeclarePairedDelimiterX\Prn[2]{(}{)}{\mspace{2mu}{#1}\;\delimsize|\;{#2}\mspace{2mu}}
\DeclarePairedDelimiterX\Bra[2]{[}{]}{\mspace{2mu}{#1}\;\delimsize|\;{#2}\mspace{2mu}}

\newcommand{\R}{\mathbb R}

\newcommand{\1}{\mathbf 1}

\renewcommand{\epsilon}{\varepsilon}

\NewDocumentCommand{\exsub}{s m O{} m}{%
  \IfBooleanT{#1}{\EE_{#2}\nolimits\bra*{#4}}%
  \IfBooleanF{#1}{\EE_{#2}\nolimits\bra[#3]{#4}}%
}

\usepackage{thmtools,thm-restate}
\usepackage{pgfplots}
\pgfplotsset{compat=newest}

\usepackage{shadethm}
\usepackage{thmdef-shaded}
\makeatletter
\define@key{thmdef}{shaded}[{}]{%
  \addtotheorempreheadhook[\thmt@envname]{%
      \setlength\shadedtextwidth{\linewidth}%
      \kvsetkeys{thmt@shade}{#1}\vspace{\thmt@style@spaceabove}\begin{shadebox}}%
    \addtotheorempostfoothook[\thmt@envname]{\end{shadebox}\vspace{\thmt@style@spacebelow}}%
  }
\makeatother

\declaretheoremstyle[
]{thmsty}

\declaretheorem[
  name=Theorem,
  refname={Theorem,Theorems},
  style=thmsty,
]{theorem}

\declaretheorem[
  name=Lemma,
  refname={Lemma,Lemmas},
  style=thmsty,
]{lemma}

\declaretheorem[
  name=Assumption,
  refname={Assumption,Assumptions},
  style=thmsty,
]{assumption}

\declaretheorem[
  name=Example,
  refname={Example,Examples},
  style=thmsty,
]{example}

\crefname{algorithm}{Algorithm}{Algorithms}
\crefname{line}{Line}{Lines}
\crefname{section}{Section}{Sections}
\crefname{appendix}{Appendix}{Appendices}
\crefname{table}{Table}{Tables}
\crefname{figure}{Figure}{Figures}
\crefname{equation}{}{}
\Crefname{equation}{Eq.}{Eqs.}

\usepackage{newfloat}
\usepackage{listings}
\DeclareCaptionStyle{ruled}{labelfont=normalfont,labelsep=colon,strut=off} 
\floatstyle{ruled}
\newfloat{listing}{tb}{lst}{}
\floatname{listing}{Listing}
\title{Optimal Transport with Cyclic Symmetry}
\author{
    Shoichiro Takeda,
    Yasunori Akagi,
    Naoki Marumo,
    Kenta Niwa
}
\affiliations{
    NTT Corporation,\\
    \{shoichiro.takeda, yasunori.akagi, naoki.marumo, kenta.niwa\}@ntt.com
}

\usepackage{bibentry}

\begin{document}

\maketitle

\begin{abstract}
We propose novel fast algorithms for optimal transport (OT) utilizing a cyclic symmetry structure of input data. Such OT with cyclic symmetry appears universally in various real-world examples: image processing, urban planning, and graph processing. Our main idea is to reduce OT to a small optimization problem that has significantly fewer variables by utilizing cyclic symmetry and various optimization techniques. On the basis of this reduction, our algorithms solve the small optimization problem instead of the original OT. As a result, our algorithms obtain the optimal solution and the objective function value of the original OT faster than solving the original OT directly. In this paper, our focus is on two crucial OT formulations: the linear programming OT (LOT) and the strongly convex-regularized OT, which includes the well-known entropy-regularized OT (EROT). Experiments show the effectiveness of our algorithms for LOT and EROT in synthetic/real-world data that has a strict/approximate cyclic symmetry structure. Through theoretical and experimental results, this paper successfully introduces the concept of symmetry into the OT research field for the first time.
\end{abstract}

\section{Introduction}\label{sec:introduction}
Given two probability vectors and a cost matrix, the discrete optimal transport (OT) problem seeks an optimal solution to minimize the cost of transporting the probability vector toward another one.
Its total transportation cost is an effective tool that compares two probability vectors.
Therefore, OT has been studied in various research areas, e.g., text embedding~\cite{kusner2015wmd}, image matching~\cite{liu2020semanticcorrespondence}, domain adaptation~\cite{courty2017domainadaptation}, graph comparison~\cite{nikolentzos2017graph}, and interpolation~\cite{solomon2015convWD}.

There are many formulations for OT.
\citet{kantorovich1942transfer} was the first to formulate OT as the linear programming problem, and the linear OT (LOT) made great progress toward solving OT.
Recently, the strongly convex-regularized OT (SROT) has attracted much attention, especially, the entropy-regularized OT (EROT)~\cite{cuturi2013sinkhorn,blondel2018sparse,cuturi2018OTsurvey,guo2020apdc}.
SROT is superior to LOT in terms of guaranteeing a unique solution and computational stability.

Many algorithms have been studied to solve OT.
The network simplex algorithm \cite{ahuja1993network} is a well-known classical algorithm for LOT and has been widely used.
The Sinkhorn algorithm~\cite{cuturi2013sinkhorn} and primal-dual descent algorithms~\cite{dvurechensky2018apdg,guo2020apdc} have been proposed to solve EROT faster.
Recently, algorithms utilizing special structures of input data have been in the spotlight for solving OT faster, e.g., algorithms that utilize the low-rankness of the input data~\cite{tenetov2018lowrank,altschuler2019lowrank}.
Besides, several algorithms utilize the Gibbs kernel structure of the input cost matrix in the Sinkhorn algorithm, such as separability~\cite{solomon2015convWD,bonneel2016sep} and translation invariance~\cite{getreuer2013gaussconv,cuturi2018OTsurvey}.

\begin{figure*}[t]
    \begin{center}
        \includegraphics[width=0.73\linewidth]{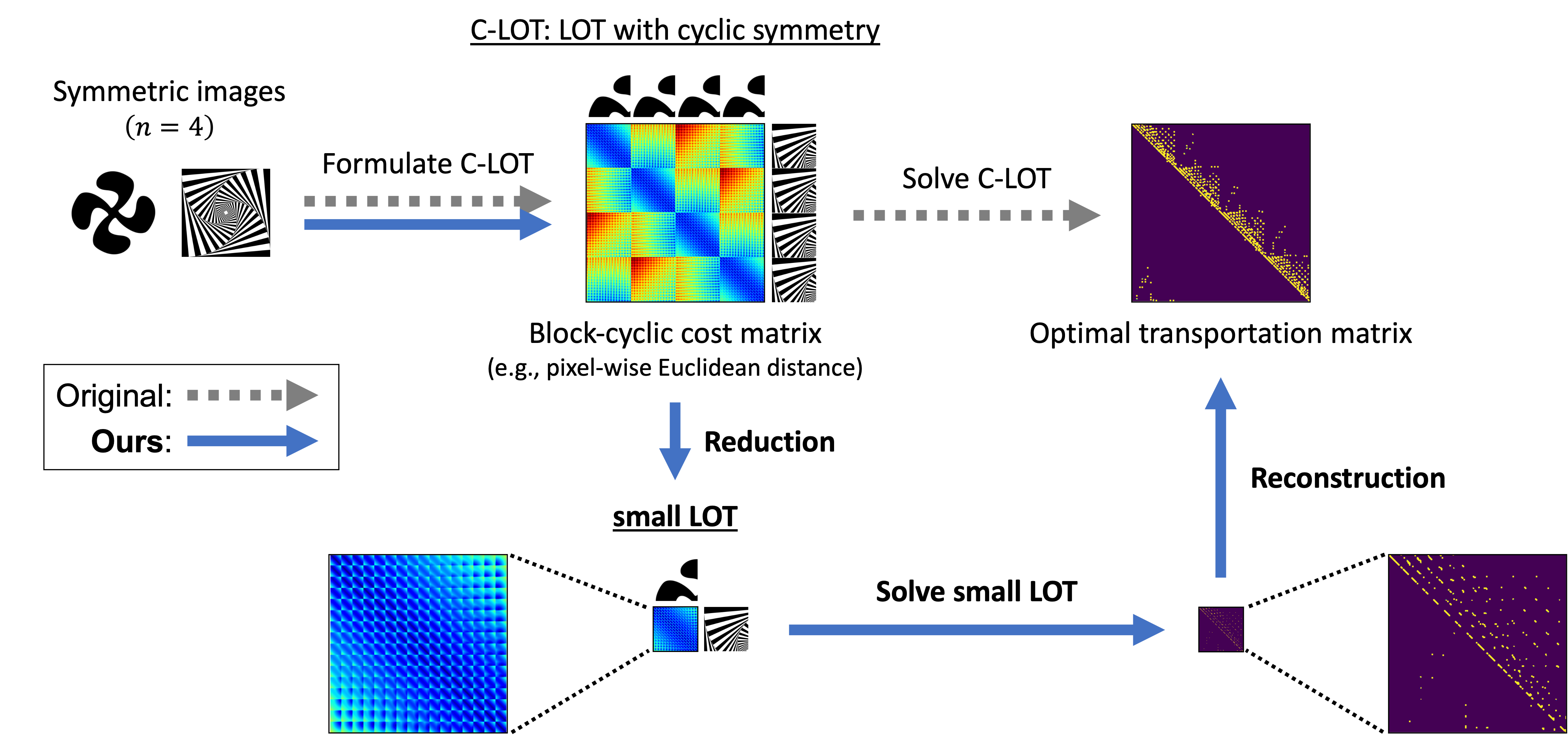}
    \end{center}
    \caption[]{
        Overview of our algorithm for LOT with cyclic symmetry (C-LOT).
        This algorithm reduces C-LOT to a small LOT that has significantly fewer variables and solves the small LOT instead, resulting in fast computation.
        Note that the small cost matrix is not just a part of the original one; it aggregates the original cost matrix on the basis of cyclic symmetry, see~\cref{eq:definition_G}.
    }
    \label{fig:overview} 
\end{figure*}

In this paper, we propose novel fast algorithms for OT utilizing a new special structure, \emph{cyclic symmetry}, of input data.
Specifically, we assume $n$-order cyclic symmetry for the input data; the input $d$-dimensional probability vector is a concatenation of $n$ copies of an $m (\coloneqq d/n)$-dimensional vector, and the input $d \times d$ cost matrix is a block-circulant matrix consisting of $n$ matrices with size $m \times m$ (see \cref{assumption:inputs}).
Such OT with cyclic symmetry appears universally in various real-world examples: image processing, urban planning, and graph processing (see examples in \cref{sec:c-rot}).
%
%
Intuitively, we can obtain an optimal solution to such a problem faster by solving OT for only one of the symmetric components of the input data and concatenating $n$ copies of the obtained solution.
However, this approach cannot work due to ignoring interactions between the symmetric components (see \cref{apsec:counter_example}).
Unlike such an intuitive way, we propose novel fast algorithms utilizing cyclic symmetry for two crucial OT formulations: LOT and SROT.

First, we propose a fast algorithm for LOT with cyclic symmetry (C-LOT).
\cref{fig:overview} shows an overview of this algorithm.
Our main idea is to reduce C-LOT, which has $d^2$ variables, to a small LOT, which has only $m^2$ variables, by utilizing cyclic symmetry.
To achieve this reduction, we introduce auxiliary variables considering cyclic symmetry and rewrite C-LOT as a $\min$-$\min$ optimization problem.
Surprisingly, the inner $\min$ problem can be solved analytically, and the $\min$-$\min$ problem becomes a small LOT.
Therefore, this algorithm solves C-LOT faster by solving the small LOT instead.
Using the network simplex algorithm to solve the small LOT, its time complexity bound becomes $O(m^3 \log m \log (m \norm*{\vect{C}}_{\infty}) + d^2)$ where $\vect{C}$ is the cost matrix and $\norm*{\vect{C}}_{\infty} \coloneqq \max_{i, j} \abs{C_{ij}}$.
This is greatly improved from $O(d^3 \log d \log (d \norm*{\vect{C}}_{\infty}))$ when solving C-LOT directly.

Next, we propose a fast algorithm for SROT with cyclic symmetry (C-SROT).
Unlike C-LOT, we cannot reduce C-SROT to a small SROT due to the regularizer.
To overcome this issue, we consider the Fenchel dual of C-SROT.
By utilizing cyclic symmetry, we show that the Fenchel dual problem has only $2m$ variables, which is significantly fewer than the $2d$ variables in the naive dual of C-SROT.
Therefore, this algorithm solves the small Fenchel dual problem by the alternating minimization algorithm \citep[Chapter~14]{amir2017altenating}.
Since the number of variables is very few, its time complexity for one iteration will be reduced, resulting in fast computation as a whole.
Especially, this algorithm for EROT with cyclic symmetry (C-EROT),  which is a subclass of C-SROT, becomes a Sinkhorn-like algorithm.
We call it cyclic Sinkhorn algorithm.
The interesting point is that the Gibbs kernel in the cyclic Sinkhorn algorithm differs from that in the original Sinkhorn algorithm and is designed by considering cyclic symmetry.
Its time complexity bound of each iteration is $O(m^2)$, which is significantly improved from $O(d^2)$ when solving C-EROT by the original Sinkhorn algorithm.

Finally, we propose a two-stage Sinkhorn algorithm for C-EROT with \emph{approximate} cyclic symmetry.
In the real world, there are many cases where the input data exhibit only approximate cyclic symmetry due to slight noise and displacement.
The cyclic Sinkhorn algorithm cannot be applied to such cases because strict cyclic symmetry of the input data is assumed.
To overcome this issue, the two-stage Sinkhorn algorithm first runs the cyclic Sinkhorn algorithm to quickly obtain a strict symmetric solution.
It then runs the original Sinkhorn algorithm to modify the solution.
As a result, this algorithm obtains the optimal solution to C-EROT with approximate cyclic symmetry faster by utilizing cyclic symmetry at the first stage.
In \cref{exp:real}, we experimentally confirmed the fast computation of this algorithm when input data have approximate cyclic symmetry.

In summary, this paper introduces the concept of symmetry into the OT research field for the first time and proposes fast cyclic symmetry-aware algorithms that solve small optimization problems instead of the original OT.
We validated the effectiveness of our algorithms in synthetic/real-world data with a strict/approximate cyclic symmetry structure.

\section{Related Work}
OT was initially formulated by \cite{monge1781memoire}.
Later \cite{kantorovich1942transfer} relaxed it as the linear programming problem, which permits splitting a mass from a single source to multiple targets.
The linear OT (LOT) is easier to solve than Monge's form and has made great progress toward solving OT.
To solve OT, many algorithms have been proposed.
For example, the network simplex algorithm \cite{ahuja1993network} is one of the classical algorithms for LOT and has been widely used.
Recently, algorithms have been proposed to solve OT faster by adding the entropy regularizer~\cite{cuturi2013sinkhorn,altschuler2017greedysinkhorn,lin2019randkhorn,alaya2019safesinkhorn}.
The dual form of the entropy-regularized OT can be solved faster by the Sinkhorn algorithm that updates dual variables via matrix-vector products~\cite{sinkhorn1967}.
For further acceleration, many improvements to the Sinkhorn algorithm have been proposed.
For example, \cite{altschuler2017greedysinkhorn}, \cite{lin2019randkhorn}, and \cite{alaya2019safesinkhorn} proposed using greedy, randomized, and safe-screening strategies, respectively, to efficiently update the dual variables.
Primal-dual algorithms have received much attention~\cite{dvurechensky2018apdg,lin2019apdm,guo2020apdc} because they report faster computation than the Sinkhorn algorithm and its variants but are rarely used in practice due to the difficulty of implementation~\cite{pham2020unbalancedOT}.
This paper focuses on the network simplex algorithm and Sinkhorn algorithm because they are widely used.

As another line of work to solve OT faster, utilizing special structures of input data has been well studied~\cite{solomon2015convWD,bonneel2016sep,cuturi2018OTsurvey,getreuer2013gaussconv,tenetov2018lowrank,altschuler2019lowrank}.
Inspired by the fact that geodesic distance matrices can be low-rank approximated~\cite{shamai2015scaling}, a low-rank approximation for the cost matrix in OT was introduced to reduce the time complexity of the Sinkhorn algorithm \cite{tenetov2018lowrank,altschuler2019lowrank}.
Several approaches have utilized the Gibbs kernel structures of the cost matrix appearing in the Sinkhorn algorithms.
The key to these approaches is to approximate the calculation involving the Gibbs kernel; by utilizing separability~\cite{solomon2015convWD,bonneel2016sep} or translation invariant~\cite{cuturi2018OTsurvey,getreuer2013gaussconv} of the Gibbs kernel on a fixed uniform grid, the matrix-vector product in the Sinkhorn algorithm can be replaced with convolutions.
Thus, it can be computed faster by, e.g., a fast Fourier transform.
This paper introduces the utilization of a new special but ubiquitous structure, cyclic symmetry, in OT.

\section{Preliminary}
\subsection{Notations}
$\R_{\geq 0}$ denotes the set of non-negative real numbers.
$\inner{\cdot}{\cdot}$ denotes the inner product; that is, for vectors $\vect{x},\vect{y} \in \R^{d}$, $\inner{\vect{x}}{\vect{y}} = \sum_{i=0}^{d-1} x_{i}y_{i}$, and for matrices $\vect{X}, \vect{Y} \in \R^{d \times d}$, $\inner{\vect{X}}{\vect{Y}} = \sum_{i,j=0}^{d-1} X_{ij} Y_{ij}$.
The probability simplex is denoted as $\Delta^{d}\coloneqq\{x_{i}\in\R^{d} \mid \sum_{i=0}^{d-1} x_{i} = 1,\,x_{i} \geq 0\}$.
$\vect{1}_{d}$ denotes the all-ones vector in $\R^{d}$.

\subsection{Regularized Optimal Transport (ROT)}
We define the regularized OT (ROT) that adds a convex regularizer to the linear OT (LOT) introduced by \cite{kantorovich1942transfer}.
Given two probability vectors $\vect{a}, \vect{b} \in \Delta^{d}$ and a cost matrix $\vect{C}\in\R^{d \times d}_{\geq 0}$, ROT can be defined as
\begin{equation}
\begin{gathered}\label{eq:problem_ROT}
    \min_{\vect{T}\in\R^{d \times d}}
    \inner{\vect{C}}{\vect{T}} + \sum_{i,j=0}^{d-1} \phi(T_{ij}),\\
    \mathrm{s.t.} 
    \quad \vect{T}\vect{1}_{d} = \vect{a},
    \quad \vect{T}^{\top}\vect{1}_{d} = \vect{b},
\end{gathered}
\end{equation}
where $\vect{T}$ is called a transportation matrix and $\phi : \R \to \R \cup \{+\infty\}$ is a convex function, called a regularizer. 
We assume $\phi(x) = +\infty$ if $x<0$; this assumption imposes the non-negative constraint on $\vect{T}$. 

ROT \cref{eq:problem_ROT} is a generalization of various OT formulations.
For example, \cref{eq:problem_ROT} leads to LOT when $\phi$ is given by
\begin{align}\label{eq:reg_for_original_OT}
    \phi(x) = 
    \begin{cases}
        0 & \mathrm{if} \  x \geq 0, \\
        +\infty & \mathrm{otherwise}.
    \end{cases}
\end{align}
Also, \cref{eq:problem_ROT} leads to the strongly convex-regularized OT (SROT) when $\phi$ is a \emph{strongly convex} function; a function $\phi$ is called strongly convex if $\phi - \frac{\mu}{2}\norm{\cdot}$ is convex for some $\mu > 0$.
As an important subclass of SROT, \cref{eq:problem_ROT} leads to the entropy-regularized OT (EROT) introduced by \cite{cuturi2013sinkhorn} when $\phi$ is given by
\begin{align}\label{eq:reg_for_entropy_ROT}
    \phi(x) = 
    \begin{cases}
        \lambda x (\log x - 1) & \mathrm{if} \  x \geq 0, \\
        +\infty & \mathrm{otherwise},
    \end{cases}
\end{align}
where $\lambda > 0$. 

\section{C-ROT: ROT with Cyclic Symmetry} \label{sec:c-rot}
This section explains our assumption of cyclic symmetry for ROT \cref{eq:problem_ROT} and real-world examples of this problem.

We assume that $\vect{a}, \vect{b}, \vect{C}$ in \cref{eq:problem_ROT} have the following $n$-order cyclic symmetry.
\begin{assumption}\label{assumption:inputs}
There exists a divisor $n$ of $d$, and the probability vectors $\vect{a},\vect{b}$ in \cref{eq:problem_ROT} have a periodic structure:
\begin{equation}
\begin{gathered}\label{eq:assumption_for_input_vectors}
    \vect{a} = \begin{pmatrix}
        \balpha \\
        \balpha \\
        \vdots \\
        \balpha \\
    \end{pmatrix},\qquad
    \vect{b} = \begin{pmatrix}
        \bbeta \\
        \bbeta \\
        \vdots \\
        \bbeta \\
    \end{pmatrix},
\end{gathered}
\end{equation}
where $\balpha, \bbeta \in \R^{m}_{\geq 0}$ and $m \coloneqq \frac{d}{n}$ is an integer.
Also, the cost matrix $\vect{C}$ in \cref{eq:problem_ROT} has a block-circulant structure:
\begin{align}\label{eq:blkstr_cost}
    \vect{C} = \begin{pmatrix}
       \vect{C}_{0} & \vect{C}_{1} & \cdots & \vect{C}_{n-1} \\
       \vect{C}_{n-1} & \vect{C}_{0} & \ddots & \vdots \\
       \vdots & \ddots & \ddots & \vect{C}_{1} \\
       \vect{C}_{1} & \cdots & \vect{C}_{n-1} & \vect{C}_{0} \\
   \end{pmatrix},
\end{align}
where $\vect{C}_{0},\dots,\vect{C}_{n-1} \in \mathbb{R}^{m \times m}_{\geq 0}$.
\end{assumption}

In this paper, we call ROT \cref{eq:problem_ROT} with \cref{assumption:inputs} \emph{Cyclic ROT (C-ROT)}.
This problem appears universally in various real-world examples given below.
\begin{example}[Image with Cyclic Symmetry]\label{example:image}
    Cyclic symmetry in images has been a central image research topic.
    Especially, because image data are represented in a rectangle form, mirror or $90^\circ$ rotational symmetry has been utilized for various tasks; mirror symmetry has been utilized for the face recognition~\cite{zhao2003facerecog} and rendering~\cite{wu2023symface}, and $90^\circ$ rotational symmetry in medical and galaxy images has been utilized for the image segmentation~\cite{shuchao2022medical_sym} and morphology prediction~\cite{dieleman2015galaxy}.
    Thus, we here consider ROT between images with cyclic symmetry, $\vect{A}$ and $\vect{B} \in \mathbb{R}^{h \times w}_{\geq 0}$.
    For images with mirror symmetry, we assume mirror symmetry along the vertical axis; 
    \begin{align}
        A_{ij} = A_{i, w - j - 1},\quad B_{ij} = B_{i, w - j - 1},  
    \end{align}
    for $0 \leq i < h$ and $0 \leq j < w$. 
    We vectorize these images by appropriately ordering pixels as follows:
    \begin{align}\label{eq:vectorization_mirror}
        \vect{a} &= \prn*{A_{\pi(0)}, A_{\pi(1)}, \ldots, A_{\pi(hw-1)}}^{\top}, \\
        \vect{b} &= \prn*{B_{\pi(0)}, A_{\pi(1)}, \ldots, B_{\pi(hw-1)}}^{\top}, \\
        \pi(k) &= 
        \begin{cases}
            \prn*{ k \bmod h, \floor{k / h} } & 0 \leq k < \frac{hw}{2} \\
            \prn*{ k \bmod h, \frac{3w}{2} - \floor{k / h} - 1 } & \frac{hw}{2} \leq k < hw
        \end{cases}.
    \end{align}
    By defining $\vect{C}$ as the Manhattan, Euclidean, or Chebyshev distance matrix between pixel positions, $\vect{a}, \vect{b}, \vect{C}$ satisfy \cref{assumption:inputs}; thus, C-ROT for $n = 2$ will appear.
    Similarly, by appropriately ordering pixels for $\vect{a},\vect{b}$ in the case of $90^\circ$ rotational symmetry, C-ROT for $n = 4$ will appear.
\end{example}

\begin{example}[Urban Planning with Cyclic Symmetry]\label{example:urban}
    ROT has straightforward applications in logistics and economy~\cite{kantorovich1942transfer,carlier2012OTeconomy}.
    Imagine a situation where planners design the structure of a city, this structure is simply given by two probability distributions: the distributions of residents $\vect{a}$ and services $\vect{b}$.
    In this context, the objective function value of ROT enables us to measure how close residents and services are and evaluate the city's efficiency.
    Several city structures, such as Howard's garden city~\cite{howard1965garden}, assume that residents and services are equally located along cyclic symmetry to improve quality of life.
    In such structures, $\vect{a},\vect{b}$ and $\vect{C}$, where $C_{ij}$ is given by the Euclidean distance between each resident $a_{i}$ and service $b_{j}$, satisfy \cref{assumption:inputs}; thus, C-ROT will appear.
\end{example}

\begin{example}[Graph with Cyclic Symmetry]\label{example:graph}
    Graphs are commonly used to model real-world data.
    For example, chemical molecules and crystal structures can be modeled using graphs~\cite{bonchev1991chemical,tian2018crystal}, and their graphs often exhibit cyclic symmetry~\cite{jaffe2002symmetry,ladd2014symmetry}.
    To compare two graphs, computing their distance has been well-studied and OT-based approaches have been proposed~\cite{nikolentzos2017graph,maretic2019got}.
    We here consider ROT for computing a distance between two graphs with cyclic symmetry.
    Following \cite{nikolentzos2017graph}, we represent features for the vertices of a graph as the eigenvectors of its adjacency matrix.
    Like chemical molecules and crystal structures, we assume the vertex features are equally distributed along cyclic symmetry.
    By defining $a_{i}=b_{j}\coloneqq\frac{1}{d}$ to ensure the same amount of outgoing/incoming flow from/to a vertex and $C_{ij}$ as the Manhattan, Euclidean, or Chebyshev distance in the eigenvectors' feature space, $\vect{a}, \vect{b}$, and $\vect{C}$ satisfy \cref{assumption:inputs}.
    Thus, C-ROT for two graphs will appear.
\end{example}

\section{Fast Algorithms for C-ROT}\label{sec:fast_algorithms}
In this section, we propose fast algorithms for C-ROT.
Note that several proofs are in the supplementary material.

\subsection{Block-Cyclic Structure of Optimal Solution}
We first show the following lemma.
\begin{lemma}\label{lemma:cyclic_optimal_solution}
    Under \cref{assumption:inputs}, there exists an optimal solution to \cref{eq:problem_ROT} that has the following structure:
    \begin{align}\label{eq:blkstr_optimal_solution}
        \vect{T} = \begin{pmatrix}
           \vect{T}_{0} & \vect{T}_{1} & \cdots & \vect{T}_{n-1} \\
           \vect{T}_{n-1} & \vect{T}_{0} & \ddots & \vdots \\
           \vdots & \ddots & \ddots & \vect{T}_{1} \\
           \vect{T}_{1} & \cdots & \vect{T}_{n-1} & \vect{T}_{0} \\
       \end{pmatrix},
    \end{align}
    where $\vect{T}_{0},\dots,\vect{T}_{n-1} \in \mathbb{R}^{m \times m}_{\geq 0}$.
\end{lemma}
\noindent The proof is shown in \cref{appendix:proof_lemma1}.

From \cref{assumption:inputs,lemma:cyclic_optimal_solution}, $\vect{C}$ and $\vect{T}$ have the same block-circulant structure.
Plugging \cref{eq:blkstr_cost,eq:blkstr_optimal_solution} into C-ROT \cref{eq:problem_ROT} yields the following optimization problem:
\begin{equation}\label{eq:problem_CROT}
\begin{gathered}
    \min_{\vect{T}_{0},\dots,\vect{T}_{n-1}\in\mathbb{R}^{m \times m}}\
    \sum_{k=0}^{n-1} \left\langle \vect{C}_{k}, \vect{T}_{k} \right\rangle
    + \sum_{k=0}^{n-1}\sum_{i,j=0}^{m-1} \phi(T_{ijk}) \\
    \text{s.t.}
    \quad\sum_{k=0}^{n-1}\vect{T}_{k}\vect{1}_{m} = \balpha,
    \quad\sum_{k=0}^{n-1}\vect{T}_{k}^{\top}\vect{1}_{m} = \bbeta,
\end{gathered}
\end{equation}
where $T_{ijk}$ is the $(i,j)$-th entry of $\vect{T}_{k}$.
Note that the objective function value of \cref{eq:problem_CROT} is exactly $\frac{1}{n}$ of that of \cref{eq:problem_ROT}.

\subsection{Algorithm for C-LOT}\label{sebsec:cyclicOT}
We here propose a fast algorithm for cyclic LOT (C-LOT), which is the special case of C-ROT \cref{eq:problem_ROT} where $\phi$ is given by \cref{eq:reg_for_original_OT}.
From \cref{eq:problem_CROT}, C-LOT \cref{eq:problem_ROT} can be rewritten as 
\begin{equation}\label{eq:problem_CLOT}
\begin{gathered}
    \min_{\vect{T}_{0},\dots,\vect{T}_{n-1}\in\mathbb{R}^{m \times m}_{\geq 0}}\
    \sum_{k=0}^{n-1} \left\langle \vect{C}_{k}, \vect{T}_{k} \right\rangle\\
    \text{s.t.}
    \quad\sum_{k=0}^{n-1}\vect{T}_{k}\vect{1}_{m} = \balpha,
    \quad\sum_{k=0}^{n-1}\vect{T}_{k}^{\top}\vect{1}_{m} = \bbeta.
\end{gathered}
\end{equation}
By introducing auxiliary variables $\vect{S} \coloneqq \sum_{k=0}^{n-1}\vect{T}_{k}$ and rewriting \cref{eq:problem_CLOT} for $\vect{S}$, we can show the following theorem.
\begin{theorem}\label{theorem:cyclic_OT}
    We consider a small LOT
    \begin{align}\label{eq:small_problem_CLOT}
        \min_{\vect{S}\in\mathbb{R}^{m \times m}_{\geq 0}} 
        \left\langle \vect{G},\vect{S}\right\rangle
        \quad\text{s.t.}
        \quad\vect{S}\vect{1}_{m} = \balpha,
        \quad\vect{S}^{\top}\vect{1}_{m} = \bbeta, 
    \end{align}
    where
    \begin{equation}\label{eq:definition_G}
        G_{ij} \coloneqq \min_{0 \leq k \leq n-1} C_{ijk}.
    \end{equation}
    Let $\vect{S}^*$ be an optimal solution of \cref{eq:small_problem_CLOT}. 
    Then, $\prn*{\vect{T}^*_k}_{k=0, \ldots, n-1}$ defined by \begin{align}\label{eq:optimal_solution_problem_CLOT}
        T^*_{ijk} =
        \begin{cases}
            S^*_{ij} & \mathrm{if}\ k = \min \left(\argmin_{0 \leq k \leq n-1} C_{ijk}\right),\\
            0 & \mathrm{otherwise}
        \end{cases}
    \end{align}
    is an optimal solution to \cref{eq:problem_CLOT}. 
    Also, the optimal objective function value of \cref{eq:problem_CLOT} is the same as that of \cref{eq:small_problem_CLOT}.
\end{theorem}

Note that $\mathrm{argmin}_{0 \leq k \leq n-1} C_{ijk}$ will return a set of indices if the same minimum value exists in several indices, and we can choose any one but the smallest index by $\min$.

\begin{proof}
    We fix $\vect{S} \coloneqq \sum_{k=0}^{n-1}\vect{T}_{k}$ in \cref{eq:problem_CLOT}. 
    The matrix $\vect{S}$ satisfies $\vect{S}\vect{1}_{m} = \balpha,\vect{S}^{\top}\vect{1}_{m} = \bbeta$ and we can rewrite \cref{eq:problem_CLOT} as 
    \begin{align}\label{eq:problem_CLOT_fix_S}
        \min_{
            \substack{
            \vect{S}\in\mathbb{R}^{m \times m}_{\geq 0},\\
            \vect{S}\vect{1}_{m} = \balpha,\ \vect{S}^{\top}\vect{1}_{m} = \bbeta
            }
        } 
        \left( \min_{
            \substack{
            \vect{T}_{0},\dots,\vect{T}_{n-1}\in\mathbb{R}^{m \times m}_{\geq 0},\\
            \sum_{k=0}^{n-1}\vect{T}_{k} = \vect{S}
            }
        } 
        \ \sum_{k=0}^{n-1} \left\langle \vect{C}_{k}, \vect{T}_{k} \right\rangle \right).
    \end{align}
    The inner problem can be solved analytically and independently for each $(i,j)$-th entry of $\vect{T}_0, \ldots, \vect{T}_{n-1}$; the optimal solution is given by \cref{eq:optimal_solution_problem_CLOT},
    and the optimal objective function value is $\inner*{\vect{G}}{\vect{S}}$.
    Next, we solve the outer optimization problem for $\vect{S}$. 
    Because $\vect{S} \in \R_{\geq 0}^{m \times m}, \vect{S}\vect{1}_{m} = \balpha$, $\vect{S}^{\top}\vect{1}_{m} = \bbeta$ and the objective function is $\inner*{\vect{G}}{\vect{S}}$, this optimization problem is identical with \cref{eq:small_problem_CLOT}. 
\end{proof}

\cref{theorem:cyclic_OT} indicates that C-LOT \cref{eq:problem_ROT} can be reduced to the small LOT \cref{eq:small_problem_CLOT}, which has significantly fewer $m^2$ variables than $d^2 = m^2 n^2$ variables of the original C-LOT \cref{eq:problem_ROT}.
Therefore, we will obtain the optimal solution to C-LOT \cref{eq:problem_ROT} by solving the small LOT \cref{eq:small_problem_CLOT} instead.
The proposed algorithm is summarized in \cref{alg:cyclicOT}.
Also, \cref{fig:overview} shows the overview of this algorithm.

We evaluate the time complexity of \cref{alg:cyclicOT}. 
The time complexity depends on the algorithm to solve the small LOT \cref{eq:small_problem_CLOT}.
We here use the network simplex algorithm, the most popular algorithm to solve LOT, to evaluate the time complexity.
\citet{tarjan1997dynamic} showed that the time complexity of the network simplex algorithm to solve LOT \cref{eq:problem_ROT} with the regularizer \cref{eq:reg_for_original_OT} is $O(d^3 \log d \log (d \norm*{\vect{C}}_{\infty}))$, where $\norm*{\vect{C}}_{\infty} \coloneqq \max_{i, j} \abs{C_{ij}}$.
This enables the time complexity of line 2 in \cref{alg:cyclicOT} to be bounded by $O(m^3 \log m \log (m \norm*{\vect{C}}_{\infty}))$.
Because line 1 and lines 3--7 can be conducted in $O(d^2)$ time, the total time complexity of \cref{alg:cyclicOT} is $O(m^3 \log m \log (m \norm*{\vect{C}}_{\infty}) + d^2)$. 
This is significantly improved from $O(d^3 \log d \log (d \norm*{\vect{C}}_{\infty}))$ when solving C-LOT \cref{eq:problem_ROT} directly.

\begin{algorithm}[t]
    \caption{Fast Algorithm for C-LOT}\label{alg:cyclicOT}
    \begin{algorithmic}[1]
        \Require $\vect{a}, \vect{b} \in \Delta^{d}$ and $\vect{C}\in\R^{d \times d}_{\geq 0}$ under \cref{assumption:inputs}.
        \State Compute $\vect{G}$ whose entry is given by \cref{eq:definition_G}.
        \State Compute the optimal solution $\vect{S}^*$ to \cref{eq:small_problem_CLOT}.
        \For{$i,j,k$}
            \State Compute $T_{ijk}$ by the relationship \cref{eq:optimal_solution_problem_CLOT}
        \EndFor
        \State Compute $\vect{T}$ by \cref{lemma:cyclic_optimal_solution} with $(T_{ijk})$
        \State \Return $\vect{T}$
    \end{algorithmic}
\end{algorithm}

\subsection{Algorithm for C-SROT} \label{subsec:algorithms_for_strongly_convex_regularizer}
We propose a fast algorithm for cyclic SROT (C-SROT) which is the special case of C-ROT \cref{eq:problem_ROT} where $\phi$ is a strongly convex regularizer.
Note that because $\phi$ defined by \cref{eq:reg_for_original_OT} is not strongly convex, we cannot apply this algorithm to C-LOT.

The following theorem follows from Fenchel's duality theorem and optimality conditions in convex analysis (see, e.g., \citep[Section~31]{rockafellar1970convex}).
\begin{theorem}\label{theorem:dual_cyclic_ROT}
    The Fenchel dual of the problem \cref{eq:problem_CROT} is
    \begin{align}
        \begin{aligned}\label{eq:dual_problem_cyclic_ROT}
        \max_{\vect{w}, \vect{z} \in \R^m}\ 
        &\inner{\vect{w}}{\balpha} + \inner{\vect{z}}{\bbeta}\\
        &-\sum_{k=0}^{n-1} \sum_{i, j=0}^{m-1} \phi^\star(w_i + z_j - C_{ijk}), 
        \end{aligned}
    \end{align}
    where $\phi^\star:\R \to \R \cup \set{+\infty}$ is the Fenchel conjugate of $\phi$ defined by $\phi^\star(y) \coloneqq \sup\{ yx - \phi(x) \mid x \in \R \}$.
    Also, the optimal solutions to the problem \cref{eq:problem_CROT}, $\vect{T}^*_k$, and to the problem \cref{eq:dual_problem_cyclic_ROT}, $\vect{w}^*$ and $\vect{z}^*$, have the following relationship:
    \begin{align} \label{eq:optimality_condition}
        T^*_{ijk} = \prn*{\phi^\star}'(w^*_i + z^*_j - C_{ijk}). 
    \end{align}
\end{theorem}
\noindent The proof is shown in \cref{appendix:proof_theorem2}.

Note that $\phi^\star$ is a smooth and differentiable convex function because $\phi$ is strongly convex.
\cref{theorem:dual_cyclic_ROT} indicates that we will obtain the optimal solution to C-SROT \cref{eq:problem_ROT} by solving the dual problem \cref{eq:dual_problem_cyclic_ROT} instead because we can reconstruct it by the relationship \cref{eq:optimality_condition} and \cref{lemma:cyclic_optimal_solution}.

We here propose to apply the alternating minimization algorithm \citep[Chapter~14]{amir2017altenating} to \cref{eq:dual_problem_cyclic_ROT}; we iteratively optimize the objective function of \cref{eq:dual_problem_cyclic_ROT} with respect to $\vect{w}$ while fixing $\vect{z}$, and vice versa.
When we fix $\vect{z}$, the partial derivative of the objective function with respect to $w_i$ is 
\begin{align} \label{eq:derivative_of_objective_function}
    \alpha_i - \sum_{k=0}^{n-1}\sum_{j=0}^{m-1}\prn*{\phi^\star}'(w_i + z_j - C_{ijk}), 
\end{align}
and $w_i$ is optimal if \cref{eq:derivative_of_objective_function} equals to 0.
Because \cref{eq:derivative_of_objective_function} monotonically decreases with respect to $w_i$, we can find such $w_i$ easily by, e.g., the well-known Newton's method.
This also applies to the optimization with respect to $\vect{z}$ while fixing $\vect{w}$.
The alternating minimization algorithm for a smooth convex function is guaranteed to attain fast convergence (see \citep{beck2013convergence} for more details). 

The distinguishing feature of this algorithm is treating a few dual variables.
If the alternating minimization algorithm is used for the dual problem of \cref{eq:problem_ROT} without considering cyclic symmetry, the number of dual variables is $2d = 2mn$.
In contrast, our algorithm treats only $2m$ dual variables, which is significantly reduced to $\frac{1}{n}$.
Therefore, the computational time required for one iteration in the alternating minimization will be considerably reduced.

\subsection{Algorithm for C-EROT}
We here propose a fast algorithm for cyclic EROT (C-EROT), which is the crucial special case of C-ROT \cref{eq:problem_ROT} where $\phi$ is given by \cref{eq:reg_for_entropy_ROT}.
Because \cref{eq:reg_for_entropy_ROT} is strongly convex, we can apply the cyclic-aware alternating minimization algorithm introduced in \cref{subsec:algorithms_for_strongly_convex_regularizer} to C-EROT. 

Because $\phi^\star(y) = \lambda \exp(\frac{y}{\lambda})$, \cref{eq:derivative_of_objective_function} can be written as 
\begin{align} \label{eq:derivative_of_entropic_reguralization}
    \alpha_i - \exp \prn*{\frac{w_i}{\lambda}} \sum_{j=0}^{m-1} K_{ij} \exp \prn*{\frac{z_j}{\lambda}}, 
\end{align}
where
\begin{align} \label{eq:def_Kij}
    K_{ij} \coloneqq \sum_{k=0}^{n-1} \exp \prn*{-\frac{C_{ijk}}{\lambda}}. 
\end{align}

From \cref{eq:derivative_of_entropic_reguralization}, we can get optimal $w_i$ in closed form:
\begin{equation}\label{eq:optimal_f_sinkhorn_logdomain}
    w_i
    = \lambda \prn*{
        \log \alpha_i
        - \log \prn*{
        \sum_{j=0}^{m-1} K_{ij} \exp \prn*{ \frac{z_j}{\lambda} }
        }
    }.
\end{equation}
We can rewrite \cref{eq:optimal_f_sinkhorn_logdomain} and describe the optimal $q_{j}$ as follows:
\begin{equation}\label{eq:optimal_sinkhorn}
    p_i
    = \frac{ \alpha_{i} }{ \sum_{j=0}^{m-1} K_{ij} q_j },\quad
    q_j
    = \frac{ \beta_{j} }{ \sum_{i=0}^{m-1} K_{ij} p_i },
\end{equation}
where $p_i \coloneqq \exp \prn*{ \frac{w_i}{\lambda} },\ q_j \coloneqq \exp \prn*{ \frac{z_j}{\lambda} }$.
This algorithm resembles the Sinkhorn algorithm \cite{cuturi2013sinkhorn}; we call it \emph{cyclic Sinkhorn algorithm}.
Note that the optimal solution $\vect{T}$ to C-EROT \cref{eq:problem_ROT} can be easily reconstructed from the optimal $\vect{w}$ and $\vect{z}$ by \cref{eq:optimality_condition} and \cref{lemma:cyclic_optimal_solution}.
The proposed algorithm is summarized in \cref{alg:cycleSinkhorn}.
\begin{algorithm}[t]
    \caption{Fast Cyclic Sinkhorn Algorithm for C-EROT}\label{alg:cycleSinkhorn}
    \begin{algorithmic}[1]
        \Require $\vect{a}, \vect{b} \in \Delta^{d}, \vect{C}\in\R^{d \times d}_{\geq 0}$ under \cref{assumption:inputs} and $\lambda > 0$.
        \State Compute $\vect{K}$ whose entry is given by \cref{eq:def_Kij}.
        \State Initialize $\vect{q} \gets \vect{1}_{m}$.
        \Repeat
            \State $\vect{p} \gets \balpha \oslash (\vect{K} \vect{q})$
            \Comment{$\oslash$ denotes elementwise division} \label{alg_line:a/Kq}
            \State{$\vect{q} \gets \bbeta \oslash (\vect{K}^{\top} \vect{p})$ \label{alg_line:b/Kp}}
        \Until{convergence}
        \For{$i,j,k$}
            \State $T_{ijk} \gets {p_i q_j} \exp \prn*{ - \frac{C_{ijk}}{\lambda}}$
        \EndFor
        \State Compute $\vect{T}$ by \cref{lemma:cyclic_optimal_solution} with $(T_{ijk})$
        \State \Return $\vect{T}$
    \end{algorithmic}
\end{algorithm}

We evaluate the time complexity of \cref{alg:cycleSinkhorn}.
The time complexity depends on the matrix-vector product iterations in lines 4 and 5 in \cref{alg:cycleSinkhorn} to solve the Fenchel dual problem \cref{eq:dual_problem_cyclic_ROT}.
In the original Sinkhorn algorithm, the time complexity of each iteration is $O(d^2) = O(m^2n^2)$~\cite{cuturi2013sinkhorn}.
In contrast, in our cyclic Sinkhorn algorithm, the time complexity of each iteration is $O(m^2)$; thus, our algorithm solves C-EROT significantly faster than the original Sinkhorn algorithm.

\section{Two-Stage Algorithm for C-EROT with Approximate Cyclic Symmetry} 
There are many real-world cases in which input data show only \emph{approximate} cyclic symmetry.
In \cref{example:image}, $\vect{C}$ satisfies \cref{assumption:inputs} strictly when using the pixel-wise Euclidean distance, but input distributions $\vect{a},\vect{b}$ (namely, images) often satisfy \cref{assumption:inputs} only approximately due to slight noise and displacement.
Thus, the above-proposed algorithms cannot be applied to such cases because they assume to satisfy \cref{assumption:inputs} strictly.
To overcome this issue, we here propose a fast \emph{two-stage Sinkhorn algorithm} for C-EROT with approximate cyclic symmetry.
Because EROT is commonly used thanks to its differentiability and computational efficiency \cite{cuturi2018OTsurvey,guo2020apdc}, we focused on C-EROT here.
The two-stage Sinkhorn algorithm first runs the cyclic Sinkhorn algorithm (\cref{alg:cycleSinkhorn}) to quickly obtain a strict symmetric solution.
It then runs the original Sinkhorn algorithm \cite{cuturi2013sinkhorn} to modify the solution.
Therefore, this algorithm obtains the optimal solution to C-EROT with approximate cyclic symmetry faster by utilizing cyclic symmetry at the first stage.
The proposed algorithm is described in \cref{alg:2stage_cycleSinkhorn}.
\begin{algorithm}[t]
    \caption{Fast Two-Stage Algorithm for C-EROT with Approximate Cyclic Symmetry}\label{alg:2stage_cycleSinkhorn}
    \begin{algorithmic}[1]
        \Require $\vect{a}, \vect{b} \in \Delta^{d}$, $\vect{C}\in\R^{d \times d}_{\geq 0}$ and $\lambda > 0$.
        \Statex // Stage1: Cyclic Sinkhorn algorithm 
        \For{$i = 0,\dots,m-1$}
            \State $\alpha_{i} = \frac{1}{n}\sum_{k=0}^{n-1} a_{i+mk}$
            \Comment{the average of $n$-divided $\vect{a}$}
            \State $\beta_{i} = \frac{1}{n}\sum_{k=0}^{n-1} b_{i+mk}$
            \Comment{the average of $n$-divided $\vect{b}$}
        \EndFor
        \State Compute $\vect{K}$ whose entry is given by \cref{eq:def_Kij}.
        \State Initialize $\widehat{\vect{q}} \gets \vect{1}_{m}$.
        \Repeat
            \State $\widehat{\vect{p}} \gets \balpha \oslash (\vect{K} \widehat{\vect{q}})$
            \Comment{$\oslash$ denotes elementwise division}
            \State $\widehat{\vect{q}} \gets \bbeta \oslash (\vect{K}^{\top} \widehat{\vect{p}})$
        \Until{convergence}
        \Statex // Stage2: Sinkhorn algorithm \cite{cuturi2013sinkhorn}
        \State Initialize $\vect{p},\vect{q}$ as the $n$ concatenated $\widehat{\vect{p}},\widehat{\vect{q}}$, respectively.
        \State Compute $K_{ij} = \exp \prn*{ - \frac{C_{ij}}{\lambda} }$.
        \Repeat
            \State $\vect{p} \gets \vect{a} \oslash (\vect{K} \vect{q})$
            \State $\vect{q} \gets \vect{b} \oslash (\vect{K}^{\top} \vect{p})$
        \Until{convergence}
        \State \Return $\vect{T} \gets \mathrm{diag}(\vect{p}) \vect{K} \mathrm{diag}(\vect{q})$
    \end{algorithmic}
\end{algorithm}

If satisfying \cref{assumption:inputs} strictly, the time complexity of this algorithm is the same as that of the cyclic Sinkhorn algorithm.
If not, it will be complex due to mixing the two Sinkhorn algorithms at Stages 1 and 2.
This analysis is for future research, but we experimentally confirmed that this algorithm shows fast computation when input data have approximate cyclic symmetry in \cref{exp:real}.

\section{Experiments}\label{sec:exp}
To validate the effectiveness of our algorithms, we conducted experiments on synthetic/real-world data that satisfy \cref{assumption:inputs} strictly/approximately.
In all experiments, we evaluated whether our algorithms, which solve small optimization problems instead of the original OT, show the same results as the original OT but with faster computation.
These experiments were performed on a Windows laptop with Intel Core i7-10750H CPU, 32 GB memory.
All the codes were implemented in Python.

\begin{table*}[t]
    \small
    \centering
    \caption{
        Experimental results in synthetic data.
        ``Obj. value" indicates objective function value.
    }\label{table:synthetic_results}
    \begin{tabular}[t]{cccccccc}
        \toprule
        \multirow{2}{*}{Algorithm}
        & \multirow{2}{*}{$n$}
        & \multicolumn{3}{c}{$d = 5000$}
        & \multicolumn{3}{c}{$d = 10000$} \\
        \cmidrule[0.5pt](lr){3-5}
        \cmidrule[0.5pt](lr){6-8}
        & & Obj. value & Marginal error & Time (sec.) & Obj. value & Marginal error & Time (sec.) \\
        \midrule
        Network Simplex
        & -- 
        & $6.034 \pm 0.824$ & $0.000 \pm 0.000$ & $6.523 \pm 1.013$
        & $6.526 \pm 0.917$ & $0.000 \pm 0.000$ & $33.660 \pm 3.238$\\
        \midrule
        \multirow{5}{*}{\begin{tabular}{@{}c@{}}Cyclic\\Network Simplex\end{tabular}}
        & 2 
        & $6.034 \pm 0.824$ & $0.000 \pm 0.000$ & $1.477 \pm 0.235$
        & $6.526 \pm 0.917$ & $0.000 \pm 0.000$ & $7.084 \pm 0.728$\\
        & 5 
        & $6.034 \pm 0.824$ & $0.000 \pm 0.000$ & $0.300 \pm 0.030$
        & $6.526 \pm 0.917$ & $0.000 \pm 0.000$ & $1.391 \pm 0.155$\\
        & 10 
        & $6.034 \pm 0.824$ & $0.000 \pm 0.000$ & $0.136 \pm 0.026$
        & $6.526 \pm 0.917$ & $0.000 \pm 0.000$ & $0.618 \pm 0.073$\\
        & 25 
        & $6.034 \pm 0.824$ & $0.000 \pm 0.000$ & $0.080 \pm 0.019$
        & $6.526 \pm 0.917$ & $0.000 \pm 0.000$ & $0.381 \pm 0.044$\\
        & 50 
        & $6.034 \pm 0.824$ & $0.000 \pm 0.000$ & $0.056 \pm 0.015$
        & $6.526 \pm 0.917$ & $0.000 \pm 0.000$ & $0.329 \pm 0.034$\\
        \midrule
        Sinkhorn
        & -- 
        & $6.233 \pm 0.821$ & $0.000 \pm 0.000$ & $3.271 \pm 1.445$
        & $6.745 \pm 0.916$ & $0.000 \pm 0.000$ & $14.589 \pm 4.213$\\
        \midrule
        \multirow{5}{*}{Cyclic Sinkhorn}
        & 2 
        & $6.233 \pm 0.821$ & $0.000 \pm 0.000$ & $0.918 \pm 0.463$
        & $6.745 \pm 0.916$ & $0.000 \pm 0.000$ & $3.973 \pm 0.922$\\
        & 5 
        & $6.233 \pm 0.821$ & $0.000 \pm 0.000$ & $0.207 \pm 0.170$
        & $6.745 \pm 0.916$ & $0.000 \pm 0.000$ & $1.262 \pm 0.324$\\
        & 10 
        & $6.233 \pm 0.821$ & $0.000 \pm 0.000$ & $0.116 \pm 0.036$
        & $6.745 \pm 0.916$ & $0.000 \pm 0.000$ & $0.636 \pm 0.259$\\
        & 25 
        & $6.233 \pm 0.821$ & $0.000 \pm 0.000$ & $0.093 \pm 0.036$
        & $6.745 \pm 0.916$ & $0.000 \pm 0.000$ & $0.381 \pm 0.127$\\
        & 50 
        & $6.233 \pm 0.821$ & $0.000 \pm 0.000$ & $0.067 \pm 0.034$
        & $6.745 \pm 0.916$ & $0.000 \pm 0.000$ & $0.320 \pm 0.053$\\        
        \bottomrule
    \end{tabular}
\end{table*}
\begin{table*}[t]
    \small
    \centering
    \caption{
        Experimental results in real-world data.
    }\label{table:real_results}
    \begin{tabular}[t]{cccccccc}
        \toprule
        \multirow{2}{*}{Algorithm}
        & \multicolumn{3}{c}{$(h,w) = (64,64),\quad d = 4096$}
        & \multicolumn{3}{c}{$(h,w) = (96,96),\quad d = 9216$} \\
        \cmidrule[0.5pt](lr){2-4}
        \cmidrule[0.5pt](lr){5-7}
        & Obj. value & Marginal error & Time (sec.) & Obj. value & Marginal error & Time (sec.) \\
        \midrule
        Sinkhorn
        & $4.320 \pm 2.056$ & $0.000 \pm 0.000$ & $16.610 \pm 6.502$
        & $6.296 \pm 3.100$ & $0.000 \pm 0.000$ & $117.152 \pm 53.442$\\
        \midrule
        Cyclic Sinkhorn
        & $4.289 \pm 2.048$ & $0.001 \pm 0.001$ & $3.837 \pm 1.286$
        & $6.250 \pm 3.089$ & $0.001 \pm 0.001$ & $26.087 \pm 11.985$\\
        \midrule
        Two-Stage Sinkhorn
        & $4.320 \pm 2.056$ & $0.000 \pm 0.000$ & $13.877 \pm 6.244$
        & $6.296 \pm 3.100$ & $0.000 \pm 0.000$ & $91.790 \pm 43.000$ \\
        \bottomrule
    \end{tabular}
\end{table*}
\subsection{Synthetic Data w/ Strict Cyclic Symmetry}\label{exp:syn}
We created $20$ synthetic random data for each of the two dimensions, $d \in \{5000, 10000\}$, that satisfy \cref{assumption:inputs} strictly in $n = 50$ (for details, see \cref{apsec:syn}).
For validation, we evaluated the average and standard deviation over each $20$ data of the objective function values, marginal constraint errors defined by $|| \vect{T}^{\top}\vect{1}_{d} - \vect{b} ||_{2}$, and the computation time when using different algorithms: the network simplex algorithm \cite{ahuja1993network}, \cref{alg:cyclicOT} using the network simplex algorithm in line 2 (we call it \emph{cyclic network simplex algorithm}), the Sinkhorn algorithm~\cite{cuturi2013sinkhorn}, and the cyclic Sinkhorn algorithm (\cref{alg:cycleSinkhorn}).
We set $\lambda = 0.5$ for the regularizer \cref{eq:reg_for_entropy_ROT}.
The computation time was recorded between inputting the data and outputting the optimal solution.
Because these synthetic data also satisfy \cref{assumption:inputs} for all $n$ that are divisors of $50$, namely $n \in \{2, 5, 10, 25, 50\}$, we conducted experiments for each $n$; larger $n$ leads to smaller problems that output the same result.
The network simplex algorithm was implemented using LEMON~\cite{LEMON2011}.

\cref{table:synthetic_results} lists the results.
The network simplex algorithm and cyclic one had the same optimal objective function value, but the latter showed faster computation times as $n$ becomes larger.
This was also the case when using the Sinkhorn algorithm and the cyclic one.
These results support the effectiveness of our proposed algorithms; higher cyclic symmetry (i.e., larger $n$) results in faster computation time.

\subsection{Real Data w/ Approximate Cyclic Symmetry}\label{exp:real}
For real-world data, we tested the case of mirror symmetry ($n=2$) in \cref{example:image} with the NYU Symmetry Database~\cite{NYUdataset2017}.
In this dataset, we selected $20$ images with mirror symmetry along the vertical axis (the images are shown in \cref{apsec:images}).
These images were converted to gray-scale, resized to be $64\times64$ or $96\times96$ pixels, and normalized so that the sum of the intensity is $1$.
We then obtained $\vect{a},\vect{b}$ by \cref{eq:vectorization_mirror} and $\vect{C}$ by the pixel-wise Euclidean distance.
For validation, we evaluated the same metrics as in \cref{exp:syn} over $190(={}_{20}\mathrm{C}_{2})$ image pairs.
Because EROT is commonly used in real applications \cite{cuturi2018OTsurvey}, we focused on C-EROT here and compared the Sinkhorn algorithm, the cyclic one (\cref{alg:cycleSinkhorn}), and the two-stage one (\cref{alg:2stage_cycleSinkhorn}).
Note that, in the two-stage Sinkhorn algorithm, we stopped Stage 1 before the end of convergence to prevent the solution far from the optimal one for real images (for details, see \cref{apsec:real}).
We set $\lambda$ as the same as in \cref{exp:syn} for the regularizer \cref{eq:reg_for_entropy_ROT}.

\cref{table:real_results} lists the results.
The cyclic Sinkhorn algorithm showed the fastest computation time.
However, because this algorithm assumes to satisfy \cref{assumption:inputs} strictly, its objective function value differed from that of the original Sinkhorn algorithm, and marginal error occurred.
In contrast, the two-stage Sinkhorn algorithm showed the same objective function value as that of the original one and no marginal error but with faster computation time than using the original one.
These results indicate that the cyclic Sinkhorn algorithm can be a good choice for real-world data because of its fastest computation time if users tolerate the objective function value difference and the marginal error.
If not, the two-stage Sinkhorn algorithm is promising for real-world data, which solves C-EROT with approximate cyclic symmetry faster than the original Sinkhorn algorithm.

\section{Discussions and Limitations}
Through this paper, we confirmed that our algorithms can solve C-ROT faster.
For further progress, we discuss the following future issues.
(I) In \cref{assumption:inputs}, we assume knowing the cyclic order $n$ in advance.
Because cyclic symmetry arises naturally from the physical structure of input data, this assumption is reasonable in some real-world cases.
However, we must improve our algorithms for unknown-order cyclic symmetry.
(II) It is unknown whether our algorithms can be generalized for other symmetries, e.g., dihedral symmetry \cite{gatermann2004symmetry}.
Further development of our algorithms for general symmetries remains as future work.
(III) The main contribution of this paper is showing the utilization of cyclic symmetry in OT with theoretical proofs, but we must test our algorithms in various real-world data for further development.

\section{Conclusion}
We proposed novel fast algorithms for OT with cyclic symmetry.
We showed that such OT can be reduced to a smaller optimization problem that has significantly fewer variables as higher cyclic symmetry exists in the input data.
Our algorithms solve the small problem instead of the original OT and achieve fast computation.
Through experiments, we confirmed the effectiveness of our algorithms in synthetic/real-world data with strict/approximate cyclic symmetry.
This paper cultivates a new research direction, OT with symmetry, and paves the way for future research.

\bibliography{main}

\onecolumn
\appendix
\renewcommand{\theequation}{A\arabic{equation}}
\renewcommand{\thefigure}{A\arabic{figure}}
\setcounter{equation}{0}
\setcounter{figure}{0}

\begin{center}
\textbf{\LARGE Appendices: Optimal Transport with Cyclic Symmetry}
\end{center}

\section{Simple Counter-Example to the Intuitive Utilization of Cyclic Symmetry in OT}\label{apsec:counter_example}
As explained in \cref{sec:introduction}, the intuitive way, which solves OT for only one of the symmetric components of input data and concatenates $n$ copies of the obtained solution, cannot work well.
To explain this reason clearly, we here present a simple counter-example to this intuitive utilization of cyclic symmetry in OT.

We consider the $90^\circ$ rotational symmetry case ($n=4$) of \cref{example:image} in \cref{sec:c-rot} with the following gray images.
\begin{figure*}[h]
    \begin{center}
        \includegraphics[width=0.5\linewidth]{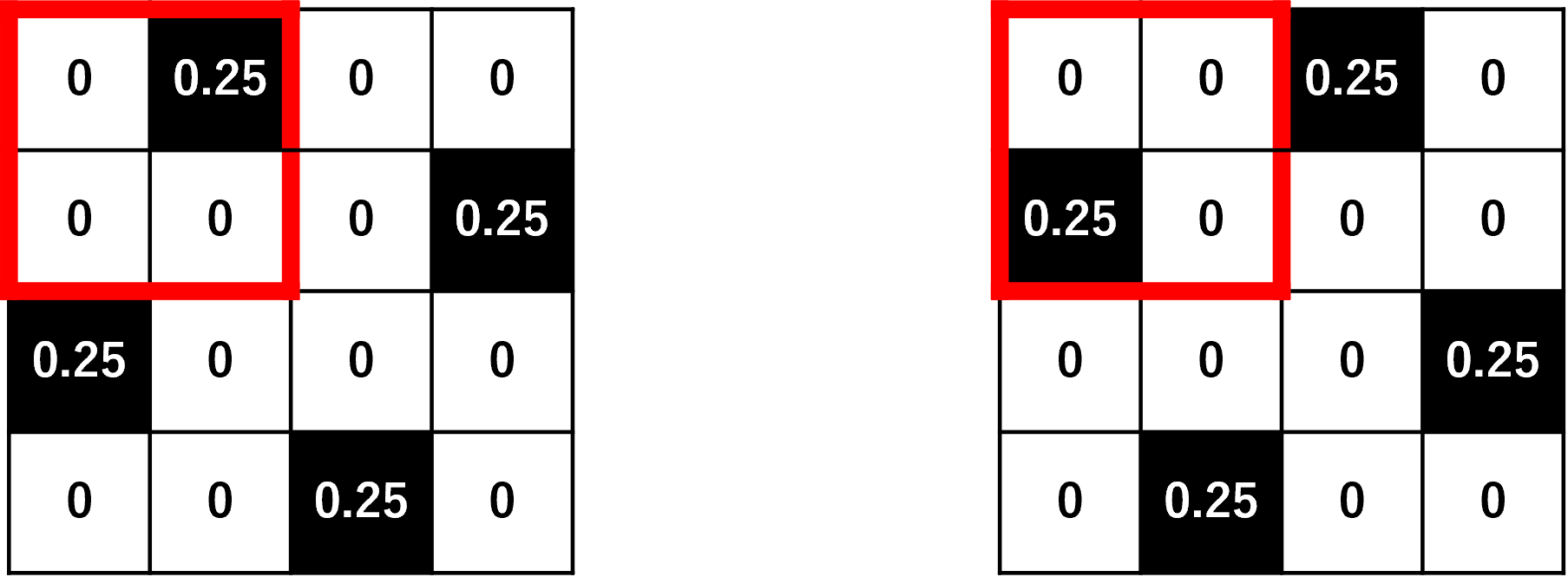}
    \end{center}
    \caption[]{
        Images with $90^\circ$ rotational symmetry ($n=4$).
    }
    \label{apfig:matrix}
\end{figure*}

\noindent These images obviously have $90^\circ$ rotational symmetry.

Intuitively, to utilize the $90^\circ$ rotational symmetry in the above images, it looks good if we consider OT for only one of the symmetric parts of the images, i.e., only the red rectangular areas in \cref{apfig:matrix}.
However, when the cost matrix $\vect{C}$ is given by the pixel-wise Euclidean distance matrix, the optimal transportation plan for the $(0,1)$-th entry (the value is $0.25$) in the left image is to be transported toward $(0,2)$-th entry (the value is $0.25$), beyond the red rectangular area, in the right image.
Thus, the OT for only one of the symmetric parts (the red rectangular areas) of the above images will give an incorrect transportation plan.

Therefore, the intuitive utilization of cyclic symmetry, i.e., solving OT for only one of the symmetric components of input data and concatenating $n$ copies of the obtained solution, cannot work well.
Consequently, we must consider the interaction between the symmetric components and develop novel fast algorithms that appropriately utilize cyclic symmetry with theoretical proofs, leading to our algorithms in the main paper.

\section{Proof of \cref{lemma:cyclic_optimal_solution}}\label{appendix:proof_lemma1}
\begin{proof}
Let $\vect{T}'$ be an optimal solution to \cref{eq:problem_ROT}. 
We define $\vect{T}^*$ as follows: 
\begin{align}\label{apeq:lemma_definition_T*}
    \vect{T}^* \coloneqq \frac{1}{n}\sum_{k=0}^{n-1} \vect{P}^{k}\vect{T}'\left(\vect{P}^{k}\right)^{\top},
\end{align}
where
\begin{align}\label{apeq:lemma_definition_P}
    \vect{P} \coloneqq \begin{pmatrix}
    \vect{O}_{m} & \vect{O}_{m} & \cdots & \vect{O}_{m} & \vect{I}_{m} \\
    \vect{I}_{m} & \vect{O}_{m} & \cdots & \vect{O}_{m} & \vect{O}_{m}\\
    \vect{O}_{m} & \ddots & \ddots & \vdots & \vdots\\
    \vdots & \ddots & \ddots & \vect{O}_{m} & \vect{O}_{m} \\
    \vect{O}_{m} & \cdots & \vect{O}_{m} & \vect{I}_{m} & \vect{O}_{m} \\
\end{pmatrix}
\end{align}
is the block-circulant permutation matrix, $\vect{I}_{m}$ denotes the $m \times m$ identity matrix, and $\vect{O}_{m}$ denotes the $m \times m$ zero matrix.

First, we will show that $\vect{T}^*$ is a feasible solution to \cref{eq:problem_ROT}.
$\vect{T}^*$ satisfies the constraints of row summation because 
\begin{align}
   \vect{T}^{*}\vect{1}_{d}
   = \frac{1}{n}\sum_{k=0}^{n-1} \vect{P}^{k}\vect{T}'\left(\vect{P}^{k}\right)^{\top}\vect{1}_{d}
   = \frac{1}{n}\sum_{k=0}^{n-1} \vect{P}^{k}\vect{T}'\vect{1}_{d}
   = \frac{1}{n}\sum_{k=0}^{n-1} \vect{P}^{k}\vect{a}
   = \frac{1}{n}\sum_{k=0}^{n-1} \vect{a} 
   = \vect{a}.
\end{align}
Similarly, we can show that $\vect{T}^*$ satisfies the constraints of column summation (i.e., $(\vect{T}^*)^{\top}\vect{1}_{d} = \vect{b}$). 
Thus, $\vect{T}^*$ is a feasible solution to \cref{eq:problem_ROT}. 

Next, we will show that $\vect{T}^*$ is an optimal solution to \cref{eq:problem_ROT}.
For this purpose, we check the optimal function value of \cref{eq:problem_ROT} when $\vect{T} = \vect{T}^*$.
Let $f(\vect{T})$ be the objective function of \cref{eq:problem_ROT}, we get
\begin{align}
    f(\vect{T}^*) = f\prn*{\frac{1}{n} \sum_{k=0}^{n-1}\vect{P}^k \vect{T}' \prn*{\vect{P}^k}^{\top}} \leq \frac{1}{n} \sum_{k=0}^{n-1} f \prn*{\vect{P}^k \vect{T}' \prn*{\vect{P}^k}^{\top}} 
    = \frac{1}{n} \sum_{k=0}^{n-1} f(\vect{T}')
    = f(\vect{T}').
\end{align}
Note that, we use the convexity of $f$ and Jensen's inequality in the inequality relationship of the above equation.
Thus, $\vect{T}^*$ is an optimal solution to \cref{eq:problem_ROT}.

Finally, for $l=0, \ldots, n-1$, we get
\begin{align}
    \vect{P}^l \vect{T}^* (\vect{P}^l)^{\top} = \frac{1}{n} \sum_{k=0}^{n-1} \vect{P}^{k+l} \vect{T}' (\vect{P}^{k+l})^{\top}
    = \frac{1}{n} \sum_{k=0}^{n-1} \vect{P}^{k} \vect{T}' (\vect{P}^k)^{\top} = \vect{T}^*.
\end{align}
Therefore, $\vect{T}^*$ has a block-circulant structure of \cref{eq:blkstr_optimal_solution}.
\end{proof}

\section{Proof of \cref{theorem:dual_cyclic_ROT}}\label{appendix:proof_theorem2}
\begin{proof} 
We rewrite \cref{eq:problem_CROT} with Lagrange multipliers $\vect{w}$ and $\vect{z}$ for the two equality constraints as follows:
\begin{equation}
\begin{aligned}\label{apeq:problem_CROT_lagrange}
    &\min_{\vect{T}_0,\dots,\vect{T}_{n-1} \in \R^{m \times m}}\
    \max_{\vect{w}, \vect{z} \in \R^m}\
    \sum_{k=0}^{n-1} \left\langle \vect{C}_{k}, \vect{T}_{k} \right\rangle
    + \sum_{k=0}^{n-1}\sum_{i,j=0}^{m-1} \phi(T_{ijk})
    +\left\langle \vect{w}, \balpha - \sum_{k=0}^{n-1}\vect{T}_{k}\vect{1}_{m} \right\rangle
    +\left\langle \vect{z}, \bbeta - \sum_{k=0}^{n-1}\vect{T}_{k}^{\top}\vect{1}_{m}\right\rangle
    \\
    =\ & \min_{\vect{T}_0,\dots,\vect{T}_{n-1} \in \R^{m \times m}}\
    \max_{\vect{w}, \vect{z} \in \R^m}\
    \inner{\vect{w}}{\balpha} + \inner{\vect{z}}{\bbeta}
    + \sum_{k=0}^{n-1} \inner*{\vect{C}_{k} - \vect{w} \1^\top - \1 \vect{z}^\top}{\vect{T}_{k}}
    + \sum_{k=0}^{n-1}\sum_{i,j=0}^{m-1} \phi(T_{ijk}).
\end{aligned}
\end{equation}
Note that Problem~\cref{eq:problem_CROT} is convex and the constraints are linear and that Slater's constraint qualification holds.
Hence, the strong duality holds (see, e.g., \citep[Section~5.2.3]{boyd_vandenberghe_2004}), and we can swap the $\min$- and $\max$-operations in \cref{apeq:problem_CROT_lagrange}:
\begin{align}\label{apeq:problem_CROT_dual}
    =\ &\max_{\vect{w}, \vect{z} \in \R^m}\
    \min_{\vect{T}_0,\dots,\vect{T}_{n-1} \in \R^{m \times m}}\
    \inner{\vect{w}}{\balpha} + \inner{\vect{z}}{\bbeta}
    + \sum_{k=0}^{n-1} \inner*{\vect{C}_{k} - \vect{w} \1^\top - \1 \vect{z}^\top}{\vect{T}_{k}}
    + \sum_{k=0}^{n-1}\sum_{i,j=0}^{m-1} \phi(T_{ijk})\\
    =\ &\max_{\vect{w}, \vect{z} \in \R^m}\ \inner{\vect{w}}{\balpha} + \inner{\vect{z}}{\bbeta} 
    -\sum_{k=0}^{n-1} \sum_{i, j=0}^{m-1} \phi^\star(w_i + z_j - C_{ijk}).
\end{align}
One of the optimality conditions is
\begin{align}
    T_{ijk}^* = (\phi^\star)'(w_i + z_j - C_{ijk}).
\end{align}
\end{proof}

\section{Further Details of Synthetic Data in \cref{exp:syn}}\label{apsec:syn}

We created synthetic data using the ``Random Generator" class in Python.
We set the random seed to $0$.
We here considered that the input data have $n(=50)$-order cyclic symmetry.

For creating the synthetic $d$-dimensional input probability vectors $\vect{a}$ and $\vect{b}$, we first sampled $\balpha$ and $\bbeta$ by $m(=\frac{d}{n})$-dimensional uniform distribution with the half-open interval $[0.0, 1.0)$.
We then created $\vect{a}$ and $\vect{b}$ by concatenating $n$ copies of $\balpha$ and $\bbeta$, respectively, like \cref{eq:assumption_for_input_vectors} and normalized them so that the sum of each is $1$.

For creating the synthetic input cost matrix $\vect{C}$, we first sampled $\vect{C}_{0},\dots,\vect{C}_{n-1}$ by $m \times m$-dimensional Gaussian distribution with the mean $3.0$ and the standard deviation $5.0$;
note that we selected these parameters to keep the same order of magnitude of metrics, namely the objective function value and the marginal error, in all experiments in \cref{sec:exp}.
We then add the absolute minimum value, namely $\left| \min_{k=0,\dots,n-1}\left( \min_{i,j=0,\dots,m-1} C_{ijk} \right) \right|$, to all entries of $\vect{C}_{0},\dots,\vect{C}_{n-1}$ to ensure their non-negativity.
After that, we created $\vect{C}$ by concatenating $n$ copies of  $\vect{C}_{0},\dots,\vect{C}_{n-1}$ like \cref{eq:blkstr_cost}.

\section{Selected Images for Real-World Data in \cref{exp:real}}\label{apsec:images}
\begin{figure*}[ht]
    \begin{center}
        \includegraphics[width=1.0\linewidth]{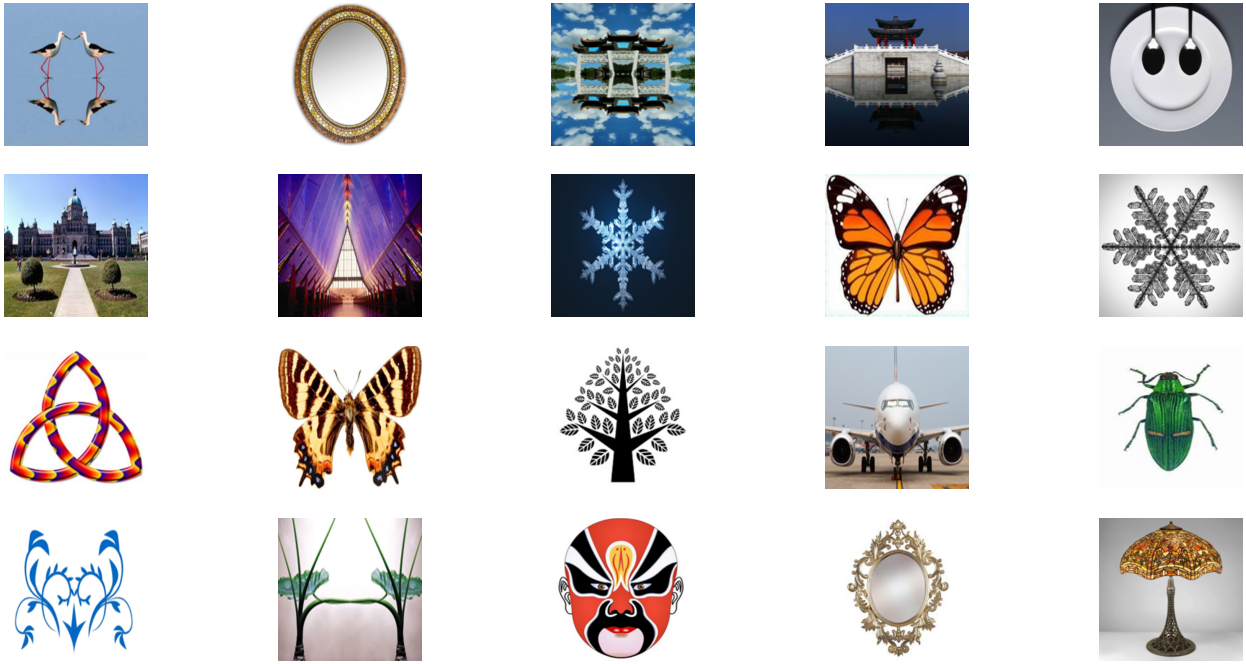}
    \end{center}
    \caption[]{
        Images with mirror symmetry, which were used in \cref{exp:real}.
        Note that we show the images after being resized to a square as in \cref{exp:real}.
    }
    \label{apfig:real_images} 
\end{figure*}

In \cref{exp:real}, we selected the 20 images with mirror symmetry ($n=2$) in the NYU Symmetry Database \cite{NYUdataset2017}.
We here show their images in \cref{apfig:real_images}.
As you see, there are various kinds of images.

\section{Further Details of the Two-Stage Sinkhorn Algorithm used in \cref{exp:real}}\label{apsec:real}
In \cref{exp:real}, we stopped Stage 1 in the two-stage Sinkhorn algorithm before convergence to prevent the solution far from optimal for real images.
For details, we first run the cyclic Sinkhorn algorithm until the marginal error $|| \left(\mathrm{diag}(\widehat{\vect{p}}) \vect{K} \mathrm{diag}(\widehat{\vect{q}})\right)^{\top}\vect{1}_{m} - \bbeta ||_{2}$ is below $1.0 \times 10^{-3}$.
We then run the original Sinkhorn algorithm until the difference between its objective function value and the value obtained by directly solving C-SROT using the original Sinkhorn algorithm is below $1.0 \times 10^{-4}$.


\def\thesection{}
\section{Appendix Reference}

\noindent
Boyd, S.; and Vandenberghe, L. 2004. \textit{Convex Optimization}. Cambridge University Press.

\end{document}